# Building extraction with vision transformer

Libo Wang, Shenghui Fang, Rui Li and Xiaoliang Meng

*Abstract*—**As an important carrier of human productive activities, the extraction of buildings is not only essential for urban dynamic monitoring but also necessary for suburban construction inspection. Nowadays, accurate building extraction from remote sensing images remains a challenge due to the complex background and diverse appearances of buildings. The convolutional neural network (CNN) based building extraction methods, although increased the accuracy significantly, are criticized for their inability for modelling global dependencies. Thus, this paper applies the Vision Transformer for building extraction. However, the actual utilization of the Vision Transformer often comes with two limitations. First, the Vision Transformer requires more GPU memory and computational costs compared to CNNs. This limitation is further magnified when encountering large-sized inputs like fine-resolution remote sensing images. Second, spatial details are not sufficiently preserved during the feature extraction of the Vision Transformer, resulting in the inability for fine-grained building segmentation. To handle these issues, we propose a novel Building Transformer (BuildFormer), with a dual-path structure. Specifically, we design a spatial-detailed context path to encode rich spatial details and a global context path to capture global dependencies. Besides, we develop a window-based linear multi-head self-attention to make the complexity of the multi-head self-attention linear with the window size, which strengthens the global context extraction by using large windows and greatly improves the potential of the Vision Transformer in processing large-sized remote sensing images. The proposed method yields state-of-the-art performance (75.74% IoU) on the Massachusetts building dataset. Code will be available.**

## I. INTRODUCTION

Building extraction using fine-resolution remote sensing images, i.e., the task of identifying building and non-building pixels in an image [1], plays a crucial role in a wide range of application scenarios such as urban planning, population statistic, economic assessment and disaster management [2-6].

Conventional methods for building extraction commonly extract hand-craft features (e.g., spectral, spatial, textural) and apply traditional machine learning methods (e.g., Support Vector Machine and Random Forest) to recognize buildings [7-9]. However, the empirically designed hand-craft features restrict the generalization ability of these traditional methods.

In the past few years, Deep Learning (DL) has become a popular approach for automatic feature learning [10] and achieved great breakthroughs in the computer vision (CV) domain [11]. In the field of remote sensing, DL methods, especially the convolutional neural network (CNN) [12], have been introduced and implemented in many geospatial tasks [13-15], especially for building extraction [16]. In comparison with conventional methods, CNN-based methods can capture various kinds of information including textures, spectrums, spatial context, and the interactions among geo-objects.

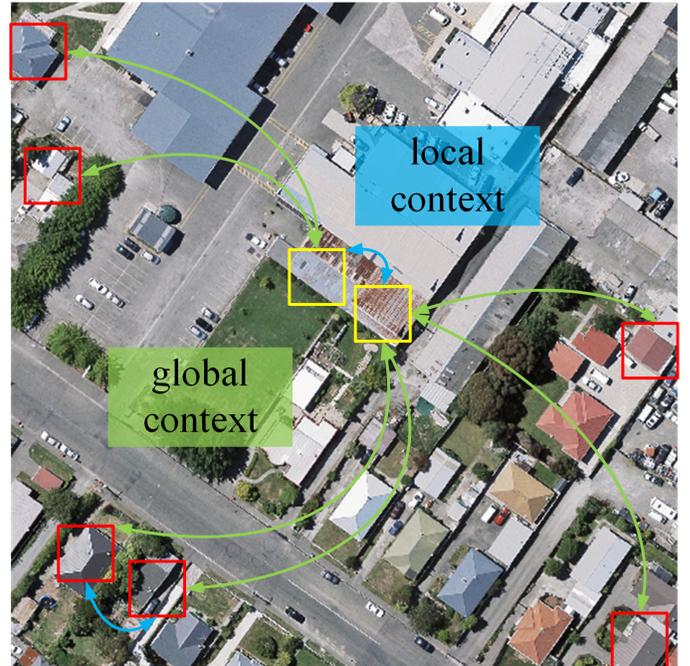

Fig. 1. Illustration of the global context and local context. The squares represent the receptive view of the convolution. The yellow regions represent the blurry building pixels where the local context is indistinguishable.

Since the pioneer CNN structure, i.e., Fully Convolutional Neural Network (FCN), was proposed for pixel-level dense prediction, a series of researches were carried out on automatic building extraction from remote sensing images [17-20]. Subsequently, the encoder-decoder structure was proposed to address the coarse-resolution segmentation of FCN-based networks by constructing a symmetrical decoder. Typical methods like UNet and SegNet restored the spatial resolution of

This work was supported in part by the National Natural Science Foundation of China (No.41971352) and Alibaba innovation research (AIR:17289315). *(Corresponding author: Libo Wang)*

L. Wang, S. Fang, and X. Meng are with School of Remote Sensing and Information Engineering, Wuhan University, Wuhan 430079, China (e-mail: wanglibo@whu.edu.cn; shfang@whu.edu.cn; xmeng@whu.edu.cn).

R. Li is with the Intelligent Control & Smart Energy (ICSE) Research Group, School of Engineering, University of Warwick, Coventry CV4 7AL, UK. (e-mail: rui.li.4@warwick.ac.uk)

extracted features progressively for fine-resolution feature representation [21, 22]. The results of these CNN-based methods, although encouraging, encounter bottlenecks in building extraction. To be specific, the CNN is designed to extract the local context and thus lacks the ability to model global context in its nature. However, the local context is often ambiguous for identifying building pixels, while the extraction will become much simpler if the global context from the whole remote sensing image is available, as illustrated in Fig.1.

For capturing the global context, the most popular way is to incorporate attention mechanisms into networks. For example, the non-local module [23], the dual attention module [24], the criss-cross attention block [25] and the object context block [26], obtained great improvements in semantic segmentation thanks to their ability in modelling global dependencies by attention mechanisms. In the field of building extraction, several attempts were made to introduce attention mechanisms for stronger feature representation, which differentiates heterogeneous buildings from complex backgrounds in fine-resolution remote sensing images [6, 27]. However, these methods still follow the CNN structure, restricting the global feature representation.

Recently, the Transformer [28], originally designed for natural language processing (NLP) tasks, comprises a hot topic in the computer vision domain, namely Vision Transformer (ViT) [29]. Different from the CNN structure, the ViT translates 2D image-based tasks into 1D sequence-based tasks. Due to the strong sequence-to-sequence modelling ability, the ViT demonstrates superior characterization of extracting global context than attention-based CNNs, obtaining numerous breakthroughs on fundamental vision tasks, such as image classification [29] and object detection [30] as well as semantic segmentation [31].

However, the actual utilization of ViTs often comes with huge memory requirements and computational costs [32, 33], which seriously affects its potential for downstream tasks like building extraction. Even though the Swin Transformer adopts the hierarchical structure and designs a window-based multi-head self-attention mechanism to improve efficiency, its complexity still increases quadratically along with the increasing size of the window [34]. Furthermore, ViTs mainly focus on capturing the global context while ignoring preserving the spatial-detailed context, but spatial details are also essential for fine-grained building segmentation in fine-resolution remote sensing images [35].

In this paper, we propose a novel Vision Transformer, namely BuildFormer, for building extraction from fine-resolution remote sensing images to address the existing issues of ViTs. Specifically, we adopt a dual-path structure to construct the BuildFormer, i.e. a global context path and a spatial-detailed context path. In the global context path, we develop a novel Transformer block to construct a Vision Transformer backbone, enhancing the ability for global context extraction. In the spatial-detailed context path, we utilize stacked convolutional layers to preserve rich spatial details. The major contributions of this paper are as follows:

1) We propose a novel Vision Transformer (BuildFormer) based on the dual-path structure, which can capture the global context while preserving spatial-detailed features.

2) We present a novel Transformer block to construct the global context path, namely BuildFormer Block (BFB), which is mainly composed of a window-based linear multi-head self-attention (W-LMHSA) and a convolutional multilayer perceptron (C-MLP).

3) The W-LMHSA reduces the complexity of the window-based multi-head self-attention (W-MHSA) [34] to linear complexity. Benefiting from this, the BuildFormer can apply larger windows to extract global features from large inputs without resulting in high computations, which is more suitable for large-scale fine-resolution remote sensing images. The C-MLP strengthens the cross-window interactions, which further enhances the ability of the BuildFormer for global information modelling.

## II. RELATED WORK

### A. CNN-based Building Extraction Methods

With the rapid development of Deep Learning, the convolutional neural network (CNN) has become the mainstream method for the automatic remote sensing building extraction task. In comparison with the conventional methods that design hand-crafted feature operators (colour, texture, shallow, etc.) [37-42] or those using active remote sensing data (LiDAR and SAR) [5, 43-46], the CNN-based methods have advantages in hierarchical feature extraction and efficiency [9, 47-50]. Although the CNN-based methods achieve many breakthroughs, their weaknesses in global information modelling limit further improvements in accuracy, as global information is crucial for detecting buildings from low-interclass and high-intraclass remote sensing images [51-53]. To address it, several studies have introduced attention mechanisms to strengthen the global feature representation for building extraction [54-57]. For example, Deng et al. [58] developed a grid-based attention gate module to extract semantic features with a global receptive field, further boosting the accuracy. Guo et al. [6] introduced the parallel attention to capturing global scene information, which further improved the accuracy of building segmentation. Pan et al. [59] combined spatial and channel attention mechanisms into the generative adversarial network and achieved advanced results. Cai et al [60] proposed a multipath hybrid attention network to enhance the performance of extracting small buildings. Since these attention-based methods relied too much on convolution operations, they failed to liberate the network from the CNN structure and have certain limitations in global information modelling.

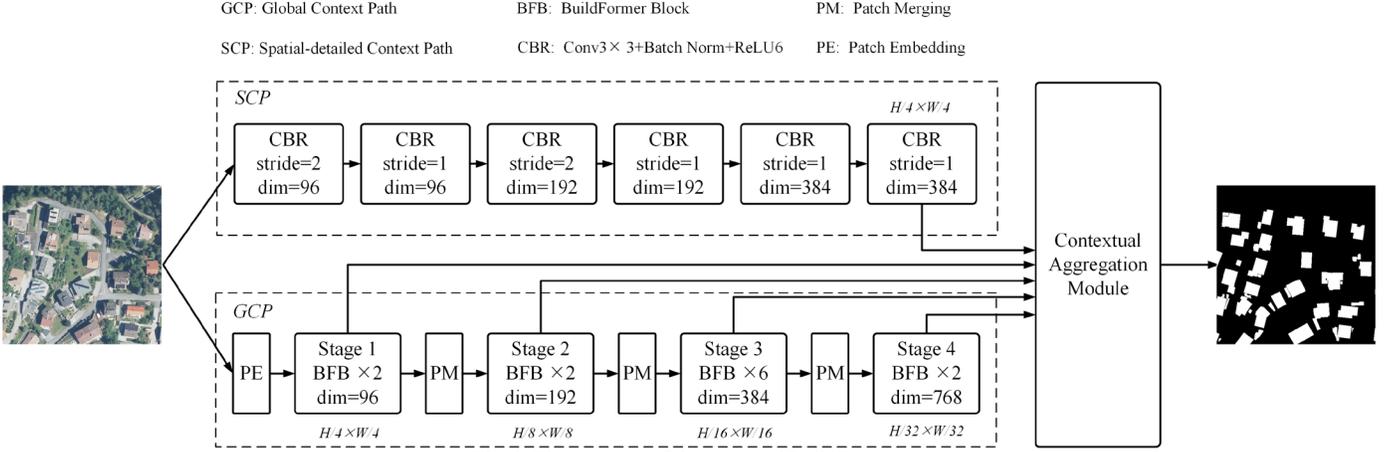

GCP: Global Context Path         BFB: BuildFormer Block         PM: Patch Merging

SCP: Spatial-detailed Context Path      CBR: Conv3×3+Batch Norm+ReLU6      PE: Patch Embedding

Fig. 2. The structure of the proposed BuildFormer.

## B. ViT-based Building Extraction Methods

ViT-based methods have brought tremendous progress and evolution for semantic segmentation [31, 61, 62]. The structure of the ViT is completely different from the CNN, which treats the 2D image as the 1D ordered sequence and applies the self-attention mechanism for global dependency modelling, demonstrating stronger global feature extraction. Driven by this, many researchers in the field of remote sensing introduced ViTs for segmentation-related tasks, such as land cover classification [63-68], urban scene parsing [69-74], change detection [75, 76], road extraction [77] and especially building extraction [78]. For example, Chen et al. [79] proposed a sparse token Transformer to learn the global dependency of tokens in both spatial and channel dimensions, achieving state-of-the-art accuracy on benchmark building extraction datasets. Yuan et al [80] introduced the widely used Swin Transformer [34] as the encoder and design a scale-adaptive decoder for multi-scale feature representation. Compared with the CNN-based methods, the global information is fully extracted by ViT-based methods. However, the spatial detailed context, meanwhile, is ignored.

## III. METHODOLOGY

### A. Overview

The structure of the proposed BuildFormer is illustrated in Fig. 2 with a Global Context Path (GCP) and a Spatial-detailed Context Path (SCP). In GCP, four BuildFormer Blocks are designed to extract four global feature maps at different scales. Meanwhile, the high-resolution spatial-detailed feature map will be generated by SCP. Finally, the four global feature maps and the spatial-detailed feature map are fed into the contextual aggregation module to generate the final semantic feature.

### B. Spatial-detailed Context Path

It is very challenging to reconcile the demand for spatial-detailed features with global dependencies simultaneously in the Vision Transformer. However, both of them are essential for obtaining high accuracy of building segmentation. To address this issue, in the proposed BuildFormer, we adopt a dual-path structure [36], which introduces a spatial-detailed context path to produce a high-resolution feature map for preserving spatial details. Concretely, we apply six (Convolution-BatchNorm-ReLU6) CBR blocks to construct this path and expand their channel dimensions progressively to encode sufficient spatial-detailed information, as shown in Fig. 2. Specifically, six standard 3×3 convolutional layers are employed and each layer is equipped with a batch normalization operation and a ReLU6 activation function. To ensure sufficient spatial details, the size of the output feature map is designed as 1/4 of the original input image.

### C. Global Context Path

The global context path is a novel self-designed Vision Transformer. The main basic modules of this path include the BuildFormer Block, Patch Embedding, and Patch Merging, as shown in Fig. 2. Due to its linear complexity, this path is more suitable for capturing global context from large-scale remote sensing images.

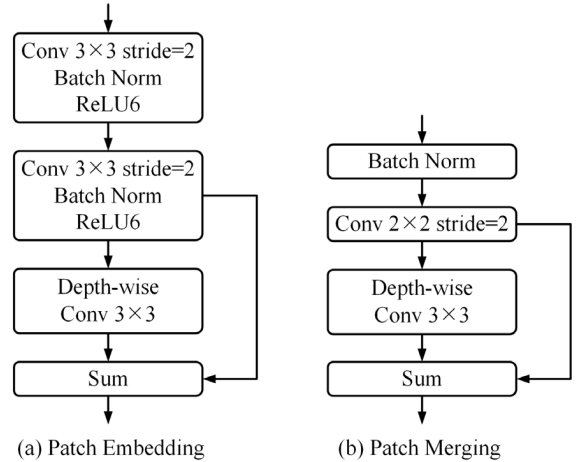

Fig. 3. (a) the Patch Embedding module, (b) the Patch Merging module.

*Patch Embedding*: The original ViT [29] utilizes linear projections to split the input image into non-overlapping patches directly. However, this scheme has limitations in modelling the structure information within patches. To overcome it, we apply convolutional layers to split the input image into overlapping patches. As shown in Fig. 3 (a), we use two 3×3 convolutional layers with a stride of 2 and a padding value of 1, while each layer is followed by a batch normalization operation and a ReLU6 activation function. Proceed by the two convolutional layers, the channel dimension of patches is expanded to 96 and the resolution is reduced to 1/4. In addition, a standard 3×3 depth-wise convolution and a residual connection are employed to enhance the relative location priors of patches.

*Patch Merging*: To obtain the hierarchical feature representation, four Patch Merging modules are employed and each module reduces the resolution of intermediate patches and expands the channel dimension. As shown in Fig. 3. (b), we first use the batch normalization operation to normalize the patches then apply a 2×2 convolutional layer to down-sampling it to 1/2 and expand its channel dimension to 2 times. Similar to the Patch Embedding module, we utilize a standard 3×3 depth-wise convolution and a residual connection to strengthen the location information extraction.

*BuildFormer Block*: Each BuildFormer Block is composed of a Window-based Linear Multi-Head Self-Attention module (W-LMHSA), a convolutional multilayer perceptron, two batch normalization operations and two residual connections, as illustrated in Fig. 4.

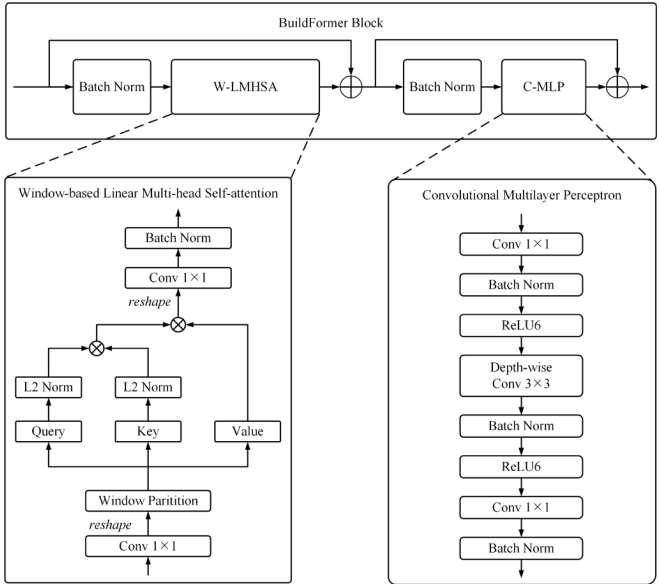

Fig. 4. Details of the BuildFormer Block.

In Swin Transformer [34], the Window-based Multi-Head Self-Attention (W-MHSA) splits the input into non-overlapping windows and performs the standard Multi-Head Self-Attention (MHSA) [28] in each local window. Benefiting from the window partition operation, the W-MHSA saves much computational burden compared to the MHSA. Even though, the computational complexity of each local window is still $O(N^2)$ due to the application of the MHSA. $N$ is the square of the window size. Thus, the W-MHSA comes with huge computations and memory requirements if using large windows.

By contrast, the proposed W-LMHSA further eliminate the high demand of W-MHSA in computations and memory based on our previous work on the linear attention mechanism [52], which makes the computational complexity linear with the window size. For each local window, the multi-head self-attention can be defined as:

$$\text{MHSA}(\boldsymbol{X}) = \text{Concat}(\text{head}_1, \ldots, \text{head}_h)\boldsymbol{W_o} \tag{1}$$

Here, $\boldsymbol{X}$ is the input vector and $h$ is the number of heads. $\boldsymbol{W_o} \in \mathbb{R}^{N \times D}$ is a projected matrix, where $D$ is the dimension of the input vector. Each head denotes a self-attention operation which can be defined as:

$$\text{Attention}(\boldsymbol{Q}, \boldsymbol{K}, \boldsymbol{V}) = \text{Softmax}_{\text{row}}\left(\frac{\boldsymbol{QK^T}}{s}\right)\boldsymbol{V} \tag{2}$$

$$\boldsymbol{Q} = \boldsymbol{X_m}\boldsymbol{W_q} \in \mathbb{R}^{N \times d} \tag{3}$$

$$\boldsymbol{K} = \boldsymbol{X_m}\boldsymbol{W_k} \in \mathbb{R}^{N \times d} \tag{4}$$

$$\boldsymbol{V} = \boldsymbol{X_m}\boldsymbol{W_v} \in \mathbb{R}^{N \times d} \tag{5}$$

where $\boldsymbol{X_m}$ is the input vector of the $m$-th head. $\boldsymbol{Q}$, $\boldsymbol{K}$ and $\boldsymbol{V}$ are the *query* feature, *key* feature and *value* feature, which are generated by the three projected matrixs $\boldsymbol{W_q}$, $\boldsymbol{W_k}$ and $\boldsymbol{W_v}$, respectively. $d$ denotes the dimension of the $m$-th head and $d=D/h$. s represents the scale factor and $s$ is set to 1 by default. $\text{Softmax}_{\text{row}}(\boldsymbol{QK^T})$ computes the similarities between each pair of pixels of the input vector and applies the softmax normalization function along each row of the similarity matrix $\boldsymbol{QK^T}$, which is the key step to model global dependencies. However, the product between $\boldsymbol{Q} \in \mathbb{R}^{N \times d}$ and $\boldsymbol{K^T} \in \mathbb{R}^{d \times N}$ belongs to $\mathbb{R}^{N \times N}$, which leads to the $O(N^2)$ computational costs and memory requirements. As $N$ is the square of the window size, the resource-demanding of the W-MHSA can increase significantly when using large windows. To address this, we simplify Eq. (2) by replacing the softmax normalization function with the first-order approximation of the Taylor expansion. Specifically, when using the softmax normalization function, the $i$-th row of the result matrix generated by Eq. (2) can be written as:

$$\text{Attention}_i(\boldsymbol{Q}, \boldsymbol{K}, \boldsymbol{V}) = \frac{\sum_{j=1}^{N} e^{\boldsymbol{q_i^T}\boldsymbol{k_j}} \boldsymbol{v_j}}{\sum_{j=1}^{N} e^{\boldsymbol{q_i^T}\boldsymbol{k_j}}} \tag{6}$$

Here, $\boldsymbol{q_i^T} \in \mathbb{R}^d$ is the $i$-th *query* feature. $\boldsymbol{k_j}$ and $\boldsymbol{v_j}$ are the $j$-th *key* feature and *value* feature, respectively. Please note that the vectors in this research are column vectors by default. Actually, Eq. (6) can be generalized to any normalization function as:

$$\text{Attention}_i(\boldsymbol{Q}, \boldsymbol{K}, \boldsymbol{V}) = \frac{\sum_{j=1}^{N} sim(\boldsymbol{q_i}, \boldsymbol{k_j}) \boldsymbol{v_j}}{\sum_{j=1}^{N} sim(\boldsymbol{q_i}, \boldsymbol{k_j})} \tag{7}$$

$$sim(\boldsymbol{q_i}, \boldsymbol{k_j}) = \phi(\boldsymbol{q_i})^T \varphi(\boldsymbol{k_j}) \tag{8}$$

$sim(\boldsymbol{q_i}, \boldsymbol{k_j})$ can measure the similarity between $\boldsymbol{q_i}$ and $\boldsymbol{k_j}$. The normalization functions $\phi(\cdot)$ and $\varphi(\cdot)$ are used to ensure $sim(\boldsymbol{q_i}, \boldsymbol{k_j}) \geq 0$. According to the first-order approximation of the Taylor expansion:

$$e^{\boldsymbol{q_i^T}\boldsymbol{k_j}} \approx 1 + \boldsymbol{q_i^T}\boldsymbol{k_j} \tag{9}$$

We set $\phi(\cdot)$ and $\varphi(\cdot)$ as the L2 normalization function to guarantee $\boldsymbol{q_i^T}\boldsymbol{k_j} \geq -1$:

$$sim(\boldsymbol{q_i}, \boldsymbol{k_j}) = 1 + \left(\frac{\boldsymbol{q_i}}{\|\boldsymbol{q_i}\|_2}\right)^T \left(\frac{\boldsymbol{k_j}}{\|\boldsymbol{k_j}\|_2}\right) \tag{10}$$

Thus, Eq. (6) can be rewritten as Eq. (11), simplified as Eq. (12) and further turned into Eq. (13):

$$\text{Attention}_i(\boldsymbol{Q}, \boldsymbol{K}, \boldsymbol{V}) = \frac{\sum_{j=1}^{N}\left(1 + \left(\frac{\boldsymbol{q}_i}{\|\boldsymbol{q}_i\|_2}\right)^T\left(\frac{\boldsymbol{k}_j}{\|\boldsymbol{k}_j\|_2}\right)\right)\boldsymbol{v}_j}{\sum_{j=1}^{N}\left(1 + \left(\frac{\boldsymbol{q}_i}{\|\boldsymbol{q}_i\|_2}\right)^T\left(\frac{\boldsymbol{k}_j}{\|\boldsymbol{k}_j\|_2}\right)\right)} \quad (11)$$

$$\text{Attention}_i(\boldsymbol{Q}, \boldsymbol{K}, \boldsymbol{V}) = \frac{\sum_{j=1}^{N}\boldsymbol{v}_j + \left(\frac{\boldsymbol{q}_i}{\|\boldsymbol{q}_i\|_2}\right)^T\sum_{j=1}^{N}\left(\frac{\boldsymbol{k}_j}{\|\boldsymbol{k}_j\|_2}\right)\boldsymbol{v}_j^T}{N + \left(\frac{\boldsymbol{q}_i}{\|\boldsymbol{q}_i\|_2}\right)^T\sum_{j=1}^{N}\left(\frac{\boldsymbol{k}_j}{\|\boldsymbol{k}_j\|_2}\right)} \quad (12)$$

$$\text{Attention}(\boldsymbol{Q}, \boldsymbol{K}, \boldsymbol{V}) = \frac{\sum_j \boldsymbol{V}_{i,j} + \left(\frac{\boldsymbol{Q}}{\|\boldsymbol{Q}\|_2}\right)\left(\left(\frac{\boldsymbol{K}}{\|\boldsymbol{K}\|_2}\right)^T\boldsymbol{V}\right)}{N + \left(\frac{\boldsymbol{Q}}{\|\boldsymbol{Q}\|_2}\right)\sum_j \left(\frac{\boldsymbol{K}}{\|\boldsymbol{K}\|_2}\right)_{i,j}^T} \quad (13)$$

Since $\sum_{j=1}^{N}\left(\frac{\boldsymbol{k}_j}{\|\boldsymbol{k}_j\|_2}\right)\boldsymbol{v}_j^T$ and $\sum_{j=1}^{N}\left(\frac{\boldsymbol{k}_j}{\|\boldsymbol{k}_j\|_2}\right)$ can be calculated and reused for each *query*, the time and memory complexity of the proposed attention based on Eq. (13) is the $O(dN)$ linear complexity.

The cross-window interaction is crucial for global dependencies modelling when using the W-MHSA. The Swin Transformer [34] introduces a shifted window operation to strengthen the cross-window interaction. This scheme, although very effective, increases the complexity of the network due to adding another shifted-window Transformer block. In this paper, we provide a convolutional multilayer perceptron (C-MLP) to strengthen the interaction within windows. In comparison with the Swin Transformer, the employment of the C-MLP can maintain competitive accuracy while improving efficiency. The detailed components of the C-MLP are illustrated in Fig. 4.

### D. Context Aggregation Module

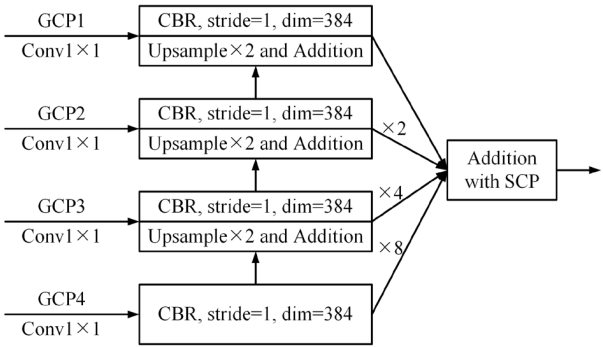

Fig. 5. Details of the Context Aggregation Module.

The output features from GCP and SCP are complementary. The feature from the SCP mainly encodes rich detailed information, while the four features generated by the GCP provide high-level global semantic information. To better fuse them, we adopt the feature fusion strategy like the feature pyramid feature (FPN) [81], as shown in Fig. 5. Specifically, the four global feature maps from the GCP are first proceeded by four $1 \times 1$ convolution layers to unify the channel dimension to 384. Then, we apply four CBR blocks as well as upsampling and addition operations to perform multi-level feature fusion. Finally, the fused global feature is further aggregated with the spatial-detailed feature from the SCP to generate the final fused feature.

### E. Loss Function

Improving the accuracy of building boundaries is vital for high-precision building extraction [35, 82-84]. Thus, we introduce the boundary supervision technology and adopt a joint loss to train the BuildFormer. The joint loss function $L$ can be defined as:

$$L = L_{ce}(Y, \hat{Y}) + L_{dice}(Y, \hat{Y}) + L_{bce}(\mathcal{L}(Y), \mathcal{L}(\hat{Y})) \quad (14)$$

where $Y$ and $\hat{Y}$ denote the predicted label and the true label, respectively. $L_{ce}$ is the cross-entropy loss. $L_{dice}$ is the dice loss. $\mathcal{L}$ represents the Laplacian convolution [85] with a kernel of $\begin{bmatrix} -1 & -1 & -1 \\ -1 & 8 & -1 \\ -1 & -1 & -1 \end{bmatrix}$ that extracts the building boundaries of the predicted label and the true label. The binary cross-entropy loss (denoted by $L_{bce}$) is employed on the extracted building boundaries.

## IV. EXPERIMENTAL SETTINGS AND DATASETS

### A. Datasets

To evaluate the performance of the proposed BuildFormer, three publicly available building datasets are considered comprehensively for conducting experiments, including the Massachusetts building dataset, WHU building dataset and Inria Aerial Image Labeling dataset. The details are as follows.

*1) Massachusetts*: The Massachusetts building dataset is composed of 151 aerial images of the Boston area with a size of $1500 \times 1500$ pixels and a ground sampling distance of 1 m. The dataset involves urban and suburban scenes, where the buildings are varied in sizes, shapes, textures and colours. Thus, this dataset is very challenging and suitable to verify the effectiveness of modules. We follow the official partition provided by the dataset and use data augmentation technologies like vertical and horizontal flip to further expand the training set. As a result, we use 411 images for training, 4 images for validation, and 10 images for testing. In the training phase, we randomly crop the images and labels into $1024 \times 1024$ pixels as the input. In the validation and testing phase, the images and labels are padded to a size of $1536 \times 1536$ pixels to ensure it is divisible by 32 (the downsampling factor of the BuildFormer). The padded parts are ignored when computing evaluation metrics.

*2) WHU*: The WHU building dataset [18] includes two types of images, i.e. satellite imagery and aerial imagery. We only use aerial images in our experiments. The aerial imagery subset covers over 450 km² and includes 22000 buildings. The spatial resolution of the RGB aerial images is 0.3 m and the size of each image is $512 \times 512$ pixels. There are 8189 image tiles in this dataset, where 4736 tiles for training, 1036 tiles for validation and 2416 tiles for testing. We follow the official partition in our experiments.

*3) Inria*: The Inria Aerial Image Labeling Dataset [86] contains 360 fine-resolution aerial images collected from five cities (Austin, Chicago, Kitsap, Tyrol and Vienna). Since the labels of the test set are publicly available, we only use the original training set in our experiments. Suggested by the official partition, the 1 to 5 tiles of each city are selected for validation and the rest for training. We first pad the original 5000×5000 images to 5120×5120 pixels, then crop them into 512×512 pixels image tiles. The image tiles, which do not contain buildings, are removed for efficient training. As a result, 9737 and 1942 image tiles are used for training and validation, respectively.

### B. Evaluation Metrics

We use the intersection over union (IoU), F1 score, precision and recall to evaluate the performance of models. These metrics are widely used in the field of building extraction [27, 35], which can be defined as follows:

$$\text{Precision} = \frac{TP}{TP + FP} \quad (15)$$

$$\text{Recall} = \frac{TP}{TP + FN} \quad (16)$$

$$\text{F1} = \frac{2 \times Precision \times Recall}{Precision + Recall} \quad (17)$$

$$\text{IoU} = \frac{TP}{TP + FN + FP} \quad (18)$$

TP, FP, and FN represent the true positive, the false positive, and the false negative, respectively.

### C. Experimental Setting

All models in the experiments were implemented with the PyTorch framework on a single NVIDIA GTX 3090 GPU with 24GB RAM. The AdamW optimizer and the cosine strategy were employed to train all models in the experiments. The random horizontal and vertical flipping were selected as data augmentation strategies. For the WHU building dataset, we trained the BuildFormer from scratch for 105 epochs. The base learning rate was set to 1e-3 and the batch size was set to 8. For the Massachachusets building dataset and the Inria Aerial Image Labelling dataset, we used BuildFormer's weight trained on the WHU building dataset, then fine-tuned it for 105 epochs with a learning rate of 5e-4. In the testing phase, we applied the data augmentation technologies like horizontal and vertical flipping, which is also known as test-time augmentation (TTA).

## V. EXPERIMENTAL RESULTS AND ANALYSIS

### A. Ablation Study

To verify the effectiveness of the proposed modules, we conducted ablation experiments on the Massachachusets building dataset.

TABLE I
THE ABLATION EXPERIMENTAL RESULTS OF SPATIAL-DETAILED CONTEXT PATH ON THE MASSACHUSETTS BUILDING DATASET.

| Method | IoU | F1 |
|---|---|---|
| BuidFormer without SCP | 74.38 | 85.30 |
| BuidFormer with SCP | 75.74 | 86.19 |

1) The effectiveness of the spatial-detailed context path (SCP): In the proposed BuildFormer, the spatial-detailed context path aims to encode rich spatial-detailed information for fine-grained building segmentation. To test its effectiveness, we remove it from BuildFormer. As listed in Table I, the utilization of the spatial-detailed context path provides an increase of 1.36% in IoU, which demonstrates its effectiveness and necessity.

Furthermore, this result also illustrates the superiority of the dual-path structure over than single-path structure for fine-grained building extraction.

2) The superiority of the global context path (GCP): The global context path in the proposed BuildFormer is a Vision Transformer backbone. To demonstrate its superiority in building extraction, we replace it with other backbones for comparison. The results show that our method yields an improvement of 2.04% in IoU compared to the Swin-Small [34] and surpassed the classical convolutional backbone ResNet101 [87] by 5.01% in IoU (Table II).

TABLE II
THE ABLATION EXPERIMENTAL RESULTS OF THE GLOBAL CONTEXT PATH ON THE MASSACHUSETTS BUILDING DATASET.

| Method | IoU | Parameter (M) |
|---|---|---|
| ResNet101 | 70.73 | 49.35 |
| Swin-Small | 73.70 | 61.54 |
| ours | 75.74 | 40.52 |

3) The effectiveness of the convolutional multilayer perceptron (C-MLP): The C-MLP aims to strengthen the cross-window interaction, improving the ability of the BuildFormer Block for capturing global context. To demonstrate its contribution to accuracy, we replace it with the standard multilayer perceptron (MLP) for ablation experiments. As illustrated in Table III, the employment of the C-MLP increases the IoU metric and the F1 score by 5.16% and 3.36%, respectively, demonstrating its effectiveness and essential.

TABLE III
THE ABLATION EXPERIMENTAL RESULTS OF SPATIAL-DETAILED CONTEXT PATH ON THE MASSACHUSETTS BUILDING DATASET.

| Method | IoU | F1 |
|---|---|---|
| BuidFormer with MLP | 70.58 | 82.75 |
| BuidFormer with C-MLP | 75.74 | 86.19 |

4) The advantages of the window-based linear multi-head self-attention (W-LMHSA): To better demonstrate the improvements of the proposed W-LMHSA, we conduct comprehensive experiments in comparison with the window-based multi-head self-attention (W-MHSA). We apply the W-LMHSA and W-MHSA to construct the BuildFormer, respectively. As shown in Table IV, the computational complexities of the W-MHSA and W-LMHSA under different window sizes are measured by the floating-point operation count (Flops) in M. The speed of the network (FPS) is measured by a 1024×1024 pixels image tile on a single NVIDIA GTX 3090 GPU. The results reveal that the proposed W-LMHSA has advantages in both accuracy and efficiency compared to the W-MHSA. Specifically, the proposed W-LMHSA can provide an improvement of 2% IoU while saving about 25% computational complexity. Besides, the W-LMHSA maintains the GPU memory requirement and the speed stable even with a large window, while the W-MHSA increases memory requirements and reduces the speed significantly.

TABLE IV
THE ABLATION STUDY OF THE W-LMHSA WITH DIFFERENT WINDOW SIZES. * MEANS THE NETWORK RUNS OUT OF MEMORY.

| Global contextual path | Window Size | Complexity (M) | Memory (MB) | Parameter (M) | Speed (FPS) | IoU |
|---|---|---|---|---|---|---|
| W-MHSA | 8 | 2.39 | 7477.36 | 40.52 | 16.73 | 72.70 |
| | 16 | 9.56 | 9301.36 | | 14.98 | 73.56 |
| | 32 | 38.24 | 16597.36 | | 10.28 | * |
| | 64 | 152.96 | * | | * | * |
| W-LMHSA (ours) | 8 | 1.77 | 7060.89 | 40.52 | 17.03 | 74.83 |
| | 16 | 7.08 | 7032.99 | | 17.17 | 75.74 |
| | 32 | 28.31 | 7024.14 | | 17.18 | 75.59 |
| | 64 | 113.25 | 7022.16 | | 17.04 | 75.36 |

### B. Comparison of State-of-the-art Methods

To further verify the effectiveness of the proposed method, we compare it with state-of-the-art methods on three publicly available datasets, i.e. the Massachusetts building dataset, WHU building dataset and Inria Aerial Image Labeling dataset. The selected methods include convolutional networks, such as U-Net [21], Deeplabv3+ [88], SRI-Net [16], DS-Net [49], BRRNet [20], SiU-Net [18], CU-Net [19], EU-Net [89], DE-Net [90], MA-FCN [48], MANet [53], MAP-Net [27], Bias-UNet [57], CBRNet [35], and ViT-based networks like SwinUperNet [34], Sparse Token Transformer (STT) [79], MSST-Net [80], BANet [72], DC-Swin [69].

For the Massachusetts building dataset, the proposed method yields a 75.74% IoU and outperforms the recent method CBRNet by 1.19% (Table V). To our knowledge base, this score is state-of-the-art on this dataset. Notably, our method achieves the highest Recall (87.52%) and surpasses other networks by a significant gap (more than 2.33%). Higher Recall means fewer building pixels missed. As shown in Fig. 6, our approach outperforms other networks in recognizing hard building pixels and maintaining the integrity of buildings, which benefits from the dual-path structure and the aggregation of the global context and spatial-detailed context.

TABLE V
QUANTITATIVE COMPARISON WITH STATE-OF-THE-ART METHODS ON THE MASSACHUSETTS BUILDING DATASET.

| Method | IoU | Precision | Recall | F1 |
|---|---|---|---|---|
| U-Net | 67.61 | 79.13 | 82.29 | 80.68 |
| DeepLab V3+ | 69.23 | 84.73 | 79.10 | 81.82 |
| MA-FCN | 73.80 | 87.07 | 82.89 | 84.93 |
| BRRNet | 73.25 | - | - | 84.56 |
| Bias-UNet | 73.49 | 83.34 | 86.15 | 84.72 |
| CBRNet | 74.55 | 86.50 | 84.36 | 85.42 |
| MANet | 70.76 | 82.00 | 83.77 | 82.88 |
| BANet | 72.20 | 83.07 | 84.66 | 83.86 |
| DC-Swin | 72.59 | 83.07 | 85.19 | 84.12 |
| BuildFormer | 75.74 | 84.90 | 87.52 | 86.19 |

TABLE VI
QUANTITATIVE COMPARISON WITH STATE-OF-THE-ART METHODS ON THE WHU BUILDING DATASET.

| Method | IoU | Precision | Recall | F1 |
|---|---|---|---|---|
| CU-Net | 87.10 | 94.60 | 91.70 | 93.13 |
| SiU-Net | 88.40 | 93.80 | 93.90 | 93.85 |
| SRI-Net | 89.23 | 95.67 | 93.69 | 94.51 |
| DE-Net | 90.12 | 95.00 | 94.60 | 94.08 |
| EU-Net | 90.56 | 94.98 | 95.10 | 95.04 |
| MA-FCN | 90.70 | 95.20 | 95.10 | 95.15 |
| MAP-Net | 90.86 | 95.62 | 94.81 | 95.21 |
| MSST-Net | 88.00 | - | - | 88.20 |
| STT | 90.48 | - | - | 94.97 |
| BuildFormer | 91.44 | 95.40 | 95.65 | 95.53 |

TABLE VII
QUANTITATIVE COMPARISON WITH STATE-OF-THE-ART METHODS ON THE INRIA AERIAL IMAGE LABELING DATASET.

| Method | IoU | Precision | Recall | F1 |
|---|---|---|---|---|
| U-Net | 70.78 | 85.18 | 80.72 | 82.89 |
| SRI-Net | 76.84 | - | - | 86.32 |
| DS-Net | 80.73 | - | - | - |
| BRRNet | 77.05 | - | - | 86.61 |
| SiU-Net | 71.40 | 84.60 | 82.10 | 83.33 |
| CBRNet | 81.10 | 89.93 | 89.20 | 89.56 |
| STT | 79.42 | - | - | 87.99 |
| SwinUperNet | 79.53 | 87.55 | 89.67 | 88.60 |
| BuildFormer | 81.44 | 88.81 | 90.75 | 89.77 |

For the WHU building dataset, the proposed method yields the best IoU (91.44%), which not only exceeds the advanced CNN-based building extraction methods by more than 0.58% but also outperforms the recent Sparse Token Transformer (STT) by 0.96% (Table VI). For the Inria Aerial Image Labeling dataset, our approach still maintains the most advanced performance with 81.44% IoU and 89.77% F1 score (Table VII). The predicted results on these two datasets are shown in Fig.7. All results reveal the importance of global context for building extraction and the superiority of dual-path structure for Vision Transformer.

## VI. CONCLUSION

In this paper, we proposed a novel Vision Transformer for building extraction from fine-resolution remote sensing images, namely the BuildFormer. Since both global context and spatial-detailed context were crucial for precise building segmentation, we designed the BuildFormer based on the dual-path structure which could capture the global information and spatial details simultaneously. Furthermore, we proposed a window-based linear multi-head self-attention to reduce the complexity of the window-based multi-head self-attention into $O(N)$. Benefiting from this, the BuildFormer could apply large windows to enhance the global context modelling without resulting in high computation. An extensive ablation study evaluated the impact of each component of the BuildFormer and experimental results on the Massachusetts, WHU, and Inria building datasets demonstrated the superiority of the proposed method in comparison with state-of-the-art methods.

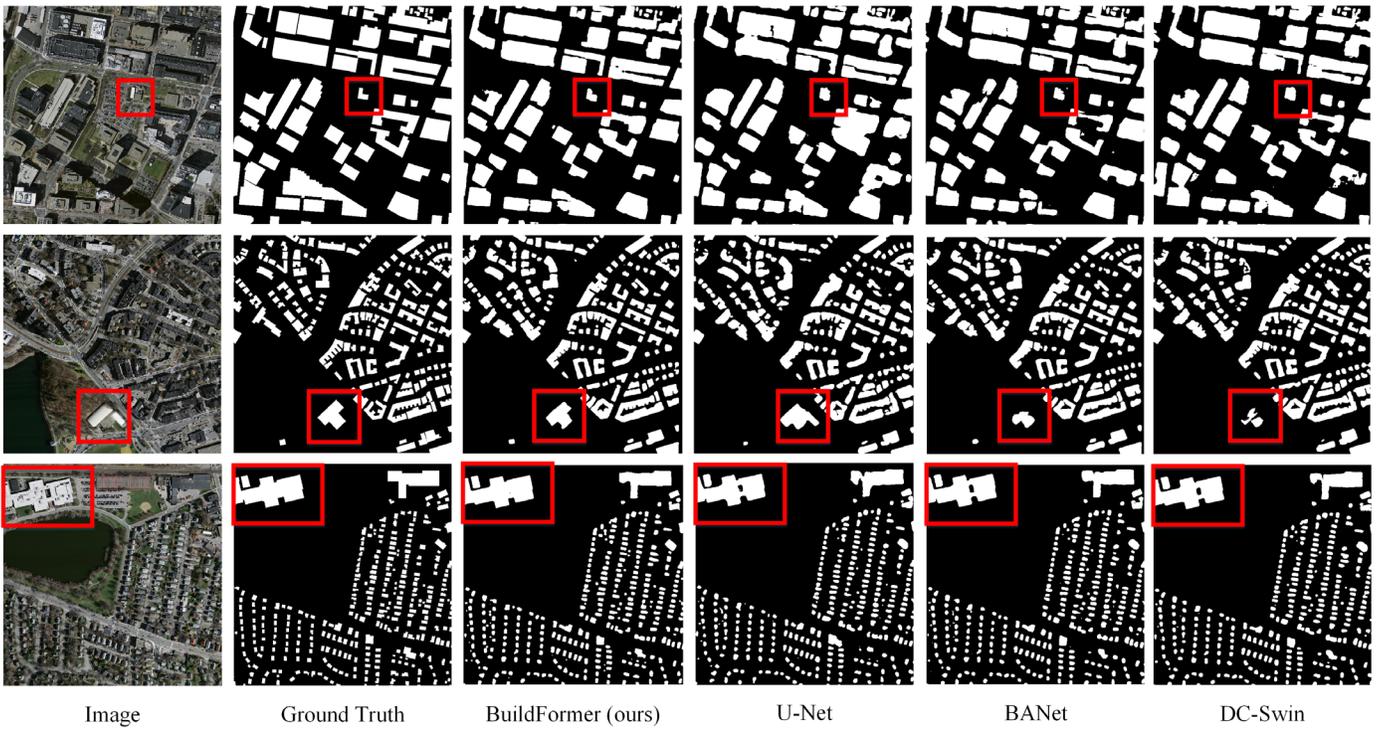

| Image | Ground Truth | BuildFormer (ours) | U-Net | BANet | DC-Swin |

Fig. 6. Visualized results of the U-Net, BANet, DC-Swin and BuildFormer (ours) on the Massachusetts Building dataset.

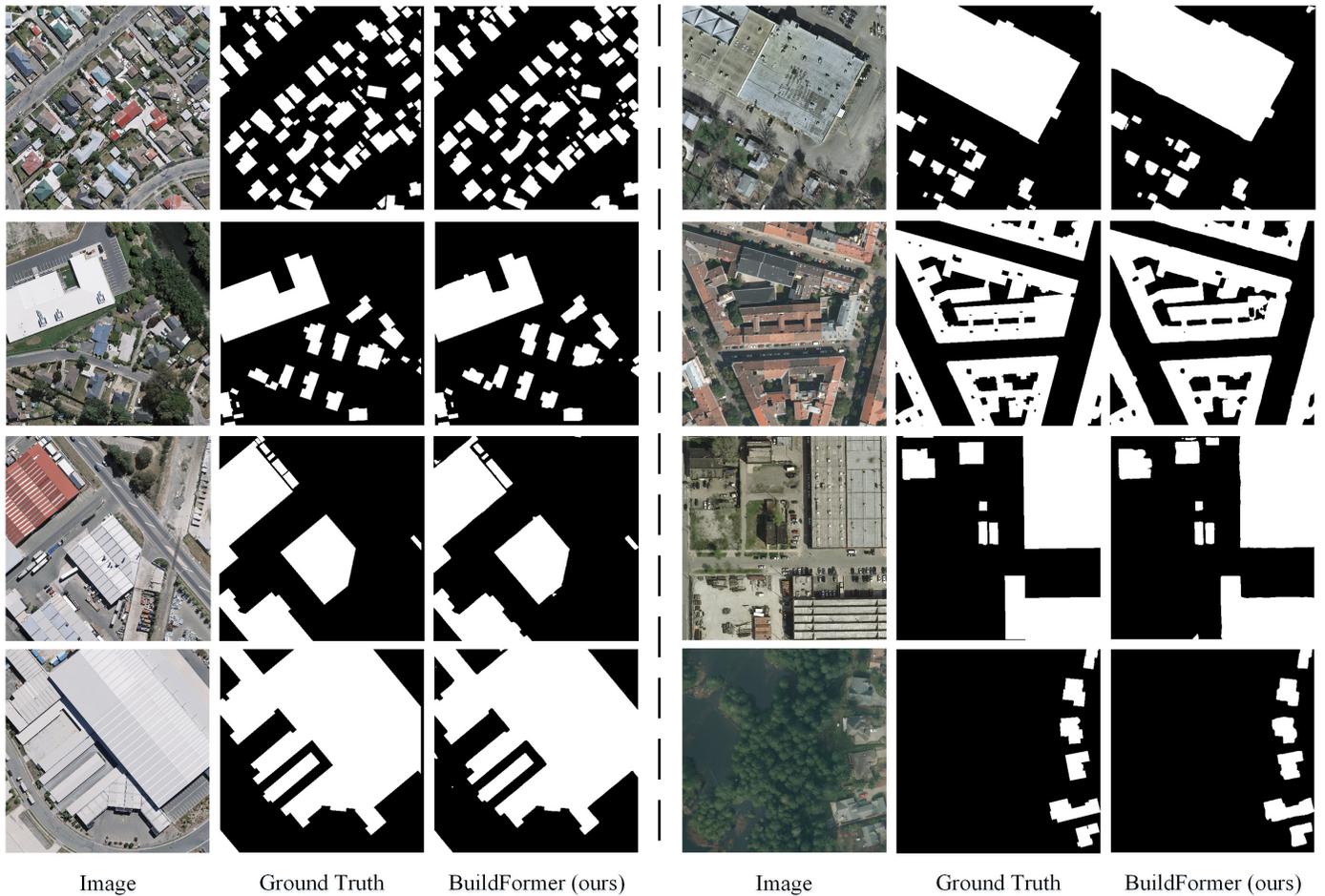

| Image | Ground Truth | BuildFormer (ours) | Image | Ground Truth | BuildFormer (ours) |

Fig. 7. Predicted results of the BuildFormer on the WHU Building dataset (left) and the Inria Aerial Image Labeling dataset (right).